\documentclass[runningheads]{llncs}
\usepackage[T1]{fontenc}
\usepackage{amssymb}
\usepackage{utfsym}
\usepackage{graphicx}%
\usepackage{multirow}%
\usepackage{amsmath,amssymb,amsfonts}%
\usepackage{mathrsfs}%
\usepackage[title]{appendix}%
\usepackage{xcolor}%
\usepackage{textcomp}%
\usepackage{manyfoot}%
\usepackage{booktabs}%
\usepackage{algorithm}%
\usepackage{algorithmic}%
\usepackage{listings}%
\usepackage[utf8]{inputenc}
\usepackage{dutchcal}
\usepackage{tabularx} 
\usepackage{array}
\usepackage{makecell}
\usepackage{enumitem}

\usepackage[misc]{ifsym}
\graphicspath{ {image/} }

\begin{document}
\title{MAPN: Enhancing Heterogeneous Sparse Graph Representation by Mamba-based Asynchronous Aggregation}
\titlerunning{MAPN}
%
\author{Xuqi Mao\inst{1,3} \and
Zhenying He\inst{1,3} \and
X. Sean Wang\inst{1,2,3}\textsuperscript{(\Letter)}}
\institute{School of Computer Science, Fudan University, Shanghai, China \and
School of Software, Fudan University, Shanghai, China \and
Shanghai Key Laboratory of Data Science, Shanghai, China\\
\email{\{xqmao17,zhenying,xywangCS\}@fudan.edu.cn}}
\maketitle              
\begin{abstract}
Graph neural networks (GNNs) have become the state of the art for various graph-related tasks and are particularly prominent in heterogeneous graphs (HetGs). However, several issues plague this paradigm: first, the difficulty in fully utilizing long-range information, known as over-squashing; second, the tendency for excessive message-passing layers to produce indistinguishable representations, referred to as over-smoothing; and finally, the inadequacy of conventional MPNNs to train effectively on large sparse graphs. To address these challenges in deep neural networks for large-scale heterogeneous graphs, this paper introduces the Mamba-based Asynchronous Propagation Network (MAPN), which enhances the representation of heterogeneous sparse graphs. MAPN consists of two primary components: node sequence generation and semantic information aggregation. Node sequences are initially generated based on meta-paths through random walks, which serve as the foundation for a spatial state model that extracts essential information from nodes at various distances. It then asynchronously aggregates semantic information across multiple hops and layers, effectively preserving unique node characteristics and mitigating issues related to deep network degradation. Extensive experiments across diverse datasets demonstrate the effectiveness of MAPN in graph embeddings for various downstream tasks underscoring its substantial benefits for graph representation in large sparse heterogeneous graphs.

\end{abstract}
\section{Introduction}

Heterogeneous Graphs (HetGs) can model numerous real-world datasets and represent diverse objects and relationships through various types of nodes and edges \cite{gamba2024exit,muppasani2024expressive,avery2024effect,agrawal2024no,liu2023generative,yang2024fine,mao2024hetfs}. The complexity and rich properties of HetGs have spurred a substantial body of research, especially in the fields of traditional graph representation learning and heterogeneous graph neural networks (HGNNs) \cite{Yu23KGTrust,Shi2022hgnn,yang23HGNAS}.

As Graph Neural Networks (GNNs) continue to evolve, a variety of models based on Message-Passing Neural Networks (MPNNs) have emerged \cite{xu2018powerful,pei2020geom}. Yet, these MPNNs face significant challenges. They often underperform on large sprse graphs where neighboring features vary greatly \cite{zheng2022graph}, and excessive message-passing layers can lead to indistinguishable embeddings, known as over-smoothing \cite{chen2020measuring}. Additionally, the problem of over-squashing has been recognized, which underscores the difficulty MPNNs have in learning from long-range neighbors \cite{topping2021understanding}.

\begin{figure*}[htbp]
\centering
\includegraphics[width=0.9\textwidth]{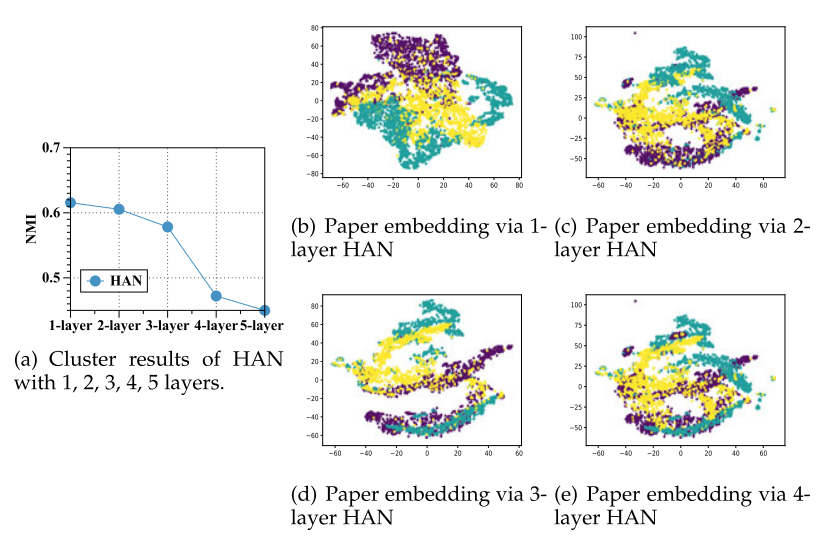}
\caption{Degration of deep neural networks.} \label{fig:MAPN_degration}
\end{figure*}

Increasing the depth of graph neural networks (GNNs) might improve the ability to capture information in graphs with more uniform node degree distributions. However, for extremely imbalanced graph structures, simply adding more layers to the network is not always an effective solution. In practice, performance degradation is often observed as the number of layers in heterogeneous GNNs increases. As illustrated in Figure~\ref{fig:MAPN_degration}, experiments on the ACM dataset show that the ability of HAN \cite{wang2019heterogeneous} to learning distinguishable paper representations diminishes significantly as the number of layers increases from one to five. This performance degradation primarily stems from the loss or dilution of information during the transmission process in deeper layers, leading to the homogenization of node features, which impairs the model’s ability to distinguish between nodes and reduces training efficiency. Simply increasing the number of layers often leads to excessive computational resource consumption and overfitting, failing to capture the true structure of the graph.

To bridge this gap, this study delves into the reasons behind model depth degradation and identifies deficiencies in current approaches to network propagation and information aggregation. We introduce the Mamba-based Asynchronous Propagation Network (MAPN), a novel approach designed to mitigate deep degradation in large, sparse heterogeneous graph representation through a state space model and a dual asynchronous aggregation mechanism. MAPN features two main components: node sequence generation and semantic information aggregation. The node sequence generation module efficiently extracts vital information from distant nodes based-on meta-path, and the semantic information aggregation module aggregates neighbor information within and across meta-paths, preserving node uniqueness during multi-hop aggregation across layers. This design ensures that the unique characteristics of each node are maintained in deep network structures, addressing the issue of deep degradation.

\section{Preliminary}\label{sec:pre}

\noindent\textbf{Definition 1} Heterogeneous Graph. A heterogeneous graph is a graph $G(V, E)$, where each node $v \in V$ is mapped to a specific node type $\theta(v) \in \mathcal{A}$ by $\theta: V \rightarrow \mathcal{A}$ and each edge $e \in E$ is mapped to a specific relation type $\kappa(e) \in \mathcal{R}$ by $\kappa: E \rightarrow \mathcal{R}$. In a heterogeneous graph, the number of distinct node types $|\mathcal{A}|$ is greater than 1, or the number of distinct relation types $|\mathcal{R}|$  is greater than 1. 
	
\noindent\textbf{Definition 2} Meta-path. A meta-path $\phi$ is a type sequence of nodes and edges defining a composite relation. It is typically represented by the notation of $A_1 \stackrel {R_1} {\longrightarrow} A_2 \stackrel {R_2} { \longrightarrow} \cdots \stackrel {R_l} {\longrightarrow} A_{l+1}$, where $A_1$ denotes the starting node type, $A_{l+1}$ denotes the ending node type, and each intermediate relation $R_i$ denotes a connection between node type $A_i$ and $A_{i+1}$. 

\noindent\textbf{Definition 3} In a simple graph $G = (V, E)$, the $l$-th layer of the Semantic Propagation Network (MPNN) can be formalized as:
\begin{equation}\label{def:MPNN}
    \begin{aligned}
    &m^{(l)}_a = \text{AGGREGATE}^{(l)}\left(\left\{H^{(l-1)}_{a^{\prime}} : a^{\prime} \in S^{(l)}(a)\right\}\right)\\
    &H^{(l)}_a = \text{UPDATE}^{(l)}\left(H^{(l-1)}_a, m^{(l)}_a\right)\\ 
    \end{aligned}
\end{equation}
where \( H^{(0)}_a \) is initialized as the feature vector \( x_a \) of node \( a \). The set \( S^{(l)}(a) \) includes the nodes from which node \( a \) directly aggregates information in the \( l \)-th layer.

\noindent\textbf{Definition 4} Synchronous MPNN. An $L$-layer MPNN is synchronous if and only if for any two layers 0 $ \leq l^\prime \leq l \leq L$ and any two nodes $u$ and $v$, there is:
\begin{equation}
    \frac{\partial H^{(l)}_a}{\partial H^{(l')}_b} \leq f_{\theta^{(L)}} \left(\frac{\partial H^{(l-1)}_{a^{\prime}}}{\partial H^{(l')}_b}\right), H^{(l-1)}_{a^{\prime}}|a^{\prime}\in\mathcal{B}_1(a).
\end{equation}
where $\theta^{(L)}$ is the parameter of the MPNN at layer $L$.

\noindent\textbf{Definition 5} State Space Model (SSM). The SSM represents dynamic systems by their states at each time step, as described by the following two equations \cite{gu2021efficiently}: $\mathbf{h}^\prime(t) = \mathbf{Ah}(t) + \mathbf{B}x(t)$ and $y(t) = \mathbf{Ch}(t) + \mathbf{D}x(t)$. Here, $\mathbf{h}(t) \in \mathbb{R}^n$ denotes the latent state of the system, $x(t) \in \mathbb{R}$ represents the input signal, $y(t) \in \mathbb{R}$ is the output signal. The parameters $\mathbf{A} \in \mathbb{R}^{n\times n}$, $\mathbf{B} \in \mathbb{R}^{n\times1}$, $\mathbf{C} \in \mathbb{R}^{m\times n}$, and $\mathbf{D} \in \mathbb{R}$ learnable. The SSM learns how to transform the input signal $x(t)$ into the latent state $\mathbf{h}(t)$, which is then used to model the system dynamics and predict its output $y(t)$.


\noindent\textbf{Problem Definition.} Given a sparse large heterogeneous graph \( G = (V, E) \) and a set of relational constraints represented by a meta-path \( \psi \), the heterogeneous graph representation learning maps each node \( a \) to a low-dimensional representation in the space defined by \( \psi \). Specifically, the goal is to generate a representation for each node through a function \( f: a \rightarrow \mathbf{a}^d, d \ll |V| \).

\section{Methodology}

\begin{figure*}[htbp]
\centering
\includegraphics[width=\textwidth]{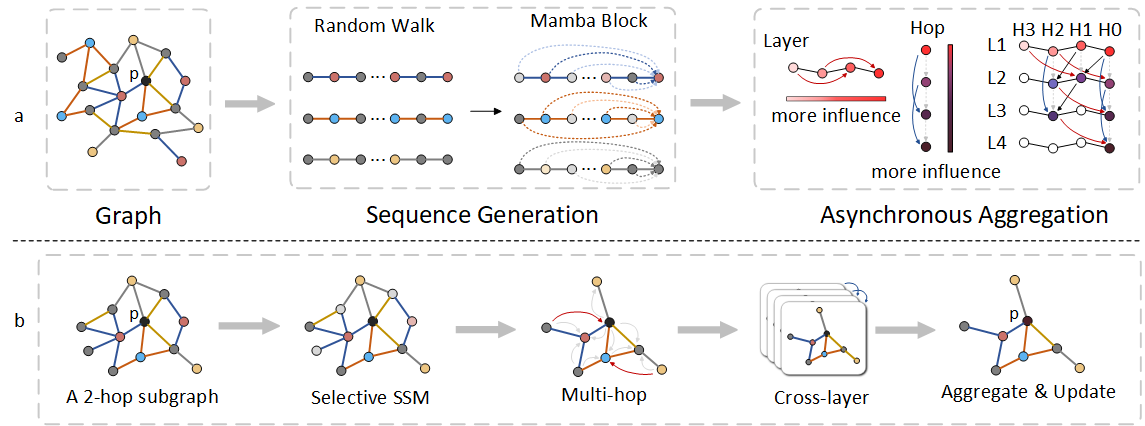}
\caption{Architecture of MAPN.} \label{fig:architecture}
\end{figure*}

This section provides a detailed description of the MAPN model architecture. As shown in Figure~\ref{fig:architecture}, MAPN first generates node sequences through random walks, and then constructs a state space model for each node based on these sequences. Using the state space model, node information is asynchronously aggregated through ``hop''-level and ``layer''-level jump connections, ultimately resulting in the node representations.

\subsection{Node Sequences Generation Module}

\subsubsection{Structure Information Extaction}

Before discussing the heterogeneous aggregation process, we first address the issue of unequal data distribution, which can negatively impact the embedding of ``hub'' nodes and inadequately represent ``cold-start'' nodes. To mitigate this, MAPN begins with MPU-based sampling using random walk with restart (RWR) in two steps: (a) Sampling Fixed-Length RWR. Beginning with a node $a \in V^\phi$, the random walker either moves to an adjacent node or returns to the starting node with probability $p$ until a fixed number of nodes for each type are collected. (b) Grouping Neighbors by Type. For each node type $A \in \mathcal{A}$, the top $k_A$ neighbors most frequently encountered during the RWR are selected as $A$-type neighbors, based on the intuition that frequently co-occurring nodes are more similar. This process enables MAPN to effectively capture and utilize the rich local information around each node, enhancing the quality of node embeddings. 

Inspired by recent advancements in Mamba, this paper introduces a sequence modeling architecture based on Mamba under meta-path \( \phi \). This approach uses a selection mechanism to preserve information from nodes at various distances, enhancing the HGNNs to capture node information in sparse large graphs. Topological data from random walks are used as input sequences, and the selection mechanism updates the hidden states based on relevant previous nodes, controlled by the current node. The variable \(g_t\), ranging from 0 to 1, allows the model to completely filter out irrelevant contexts when needed. This mechanism refines long-distance dependencies while minimizing the influence of less relevant nodes, enhancing graph attention and preserving key dependencies within long sequences. Specifically, MAPN employs a state space model based on the meta-path \( \phi \) to differentiate between relevant and irrelevant information.

\subsection{Semantic Aggregation Module}

\subsubsection{Node Information Transformation}

In heterogeneous graphs with node attributes, feature vectors for different content types of a node $a$ often have unequal dimensions and belong to distinct feature spaces. As a result, managing feature vectors of diverse dimensions within a unified framework presents significant challenges. To address this, a two-step transformation is used to capture node information.

\paragraph{Type-Specific Transformation} 

For each content type of a node, features are projected into a unified feature space via a type-specific neural network $f$. For the $n$-th content of node $a$, we have:
\begin{equation}
H_{an}=f_n\big(C_{an}\big)
\end{equation}
where $C_{an}$ is the original feature vector of $n$-th content for node $a$ and $H_{an}$ is the projected latent vector, $f_n$ is the type-specific transformation function, which can be pre-trained using different techniques for various content types. The node information of node $a$ can then be represented as:
\begin{equation}
H_a=\left\{H_{an}, n \in |C_a|\right\}
\end{equation}
where $H_a$ is the collection of latent feature vectors of node $a$, and $|C_a|$ denotes the amount of content feature of node $a$. 

\paragraph{Node Information Aggregation} 

After transforming the content for each node, all node features are standardized to the same dimension. We utilize Bidirectional LSTM, as described by \cite{hamilton2017inductive}, to aggregate the diverse set of unordered features of the node. The feature embedding process is as follows:
\begin{equation}
\hat{H}_a = \frac{\sum_{n\in H_a} \left[\overrightarrow{LSTM}\left (H_{an}\right) \bigoplus \overleftarrow{LSTM}\left (H_{an}\right)\right]}{|H_a|}
\end{equation}
where $\bigoplus$ denotes concatenation. This method enables the integration of features from neighbors of the same type, ensuring comprehensive representation in the embeddings. 

This operation aligns the projected features of all nodes to a uniform dimension, facilitating the subsequent processing of nodes of arbitrary types. 

\subsubsection{Semantics Information Aggregation}

\paragraph{Synchronous Semantic Aggregation}

Local priority by layers: Extracting distinguishable category information from the feature matrix is crucial for the success of GNNs. However, the synchronous nature of deeper GNNs leads to a loss of important original information, causing them to fail in retaining sufficient distinctions.

\begin{theorem}\label{thm:cross_layer}
    Consider an \(L\)-layer Semantic Propagation Network, suppose the activation function \(\sigma(\cdot)\) has a Lipschitz constant \(\alpha\), and the norm of each element in \(W\) is bounded, such that \(\|W\|_p \leq c\). Within a \(K\)-regular graph, for any two nodes \(a\) and \(b\), and any layer range \(0 \leq l' < l \leq L\), the following relationship is established:
    \begin{equation}
    \sum_{a \in V} \left\lvert\left\lvert\frac{\partial H^{(l)}_a}{\partial H^{(l')}_b}\right\rvert\right\rvert^2_p \leq C \sum_{a \in V} \left(\frac{\partial H^{(l-1)}_a}{\partial H^{(l')}_b}\right)^2_p, where C = \frac{\alpha^2 c^2 K^2}{(K + 1)^2}
    \end{equation}
\end{theorem}

This theorem demonstrates how influence from any layer \(l'\) propagates across \(l\) layers. For most GNN activation functions such as tanh, sigmoid, and ReLU, \(\alpha\) generally does not exceed 1. Input features \(X\) are typically normalized, keeping \(c\) small \cite{tang2023towards}. Consequently, \(C \leq 1\) in most scenarios, indicating that the impact of the original input weakens with increasing layer depth, favoring local processing. Skip connections, particularly initial ones, help maintain more original features by reducing the instances of compression, thereby avoiding excessive smoothing in deeper layers.

Local priority by hops: Information flow from higher-order neighbors must pass through lower-order ones to reach the central node. The Ollivier-Ricci curvature \( \kappa(a, b) \), which ranges from (-1, 2) \cite{bauer2011ollivier,lin2011ricci}, measures the connectivity between one-hop neighbors and helps examine the hop distance effect on information flows (IFs). Higher \( \kappa(a, b) \) indicates stronger local connectivity.  

\begin{theorem}\label{thm:multi_hop}
Consider an \(L\)-layer semantic propagation network, where the activation function \( \sigma(\cdot) \) is the identity function. Define \( \eta \) as the lower bound for the Ollivier-Ricci curvature on a \(K\)-regular graph, meaning \( \kappa(a, b) \geq \eta \) for every edge \( (a, b) \) in \( E \). If \( \eta \geq \frac{1}{2} - \frac{3}{2K} \), the following holds:
\begin{equation}
\sum_{a \in N_1(b)} \left\lvert\left\lvert\frac{\partial h_u^{(l+2)}}{\partial h_v^{(l)}}\right\rvert\right\rvert_p \geq \sum_{a \in N_2(v)} \left\lvert\left\lvert\frac{\partial H^{(l+2)}_a}{\partial H^{(l)}_v}\right\rvert\right\rvert_p    
\end{equation}
\end{theorem}

Focusing on one-hop and two-hop neighbors reveals the local structure's impact on central nodes in MAPN. The theorem shows that denser local connections (higher \( \kappa \)) and fewer one-hop neighbors (lower \( K \)) lead to greater gradient decay with hop distance, demonstrating a preference for hopping. In contrast, multi-hop propagation reduces local density (lower \( \kappa \)) and increases node degree (higher \( K \)), aiding GNNs in reducing the influence of irrelevant first-order neighbors while preserving long-distance dependencies.

\paragraph{Asychronous Semantic Aggregation}

\begin{figure}[h]
    \centering
        \includegraphics[width=0.65\textwidth]{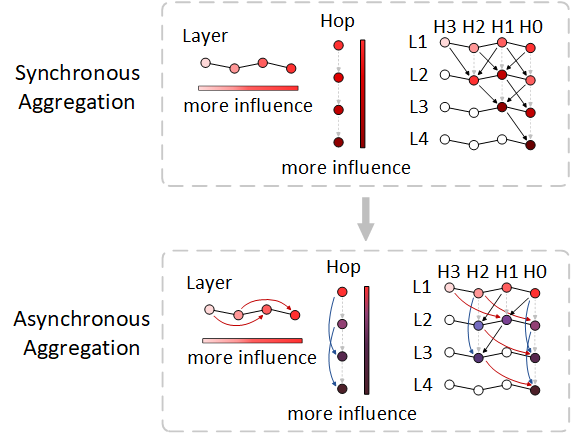}
    \caption{Asynchronous aggregation.} \label{fig:MAPN_asynchronous_aggregation}
\end{figure}

As analyzed earlier, overcoming the local priority caused by synchrony is key to enhancing MPNN performance. This section presents an asynchronous semantic propagation network that uses a multi-hop messaging method based on shortest paths to speed up communication between different node pairs, as illustrated in Fig.~\ref{fig:MAPN_asynchronous_aggregation}. This approach not only preserves original information from distant endpoints but also reduces information compression. Specifically, each node gathers information from neighbors within \(k\) hops, which is then filtered through a selective state space for aggregation, ensuring retention of distance-related information.

Intra-meta-path-aggregation: The process selectively filters and learns about the structural and semantic information of the target node, its neighbors, and their interrelationships using a selective state space. 

MAPN employs a bidirectional LSTM for aggregating node features. As described in \cite{hamilton2017inductive}, this approach captures long-distance information transmission and maintains long-term dependencies. The computation of feature embeddings is as follows:
    \begin{equation}
    \hat{H}^{\phi}(a) = \frac{\sum_{n\in H^{\phi}(a)} \left[\overrightarrow{LSTM}\left (H_{n}^{\phi}(a)\right) \bigoplus \overleftarrow{LSTM}\left (H_{n}^{\phi}(a)\right)\right]}{|H^{\phi}(a)|}
    \end{equation}
 where $\bigoplus$ denotes concatenation.
    
To unify the feature vectors of different types of nodes, MAPN aggregates the fixed-size neighbor set of each node \(a\) using LSTM. 
    \begin{equation}
    Q^{\phi}(a) = \frac{\sum_{b \in N^{\phi}(a)} \left[\overrightarrow{LSTM}\left(\hat{H}^{\phi}(b)\right) \bigoplus \overleftarrow{LSTM}\left(\hat{H}^{\psi}(b)\right)\right]}{|N^{\phi}(a)|}
    \end{equation}
where \(N^{\phi}(a)\) denotes the sampled neighbor set of node \(a\), where \(b\) is a neighbor of \(a\) within the meta-path \(\phi\).

Inter-meta-path-aggregation: Nodes across various meta-paths \(\phi\) contribute differently to the target node's representation, prompting MAPN to filter their vector representations using a selective state space \cite{gu2023mamba}:
    \begin{equation}
    \begin{aligned}
    &\alpha^{\phi}(a,b) = \frac{exp{\left(LeakyReLU\left(u^T[{Q}^{\phi}(a) \bigoplus {Q}^{\psi}(b)]\right)\right)}}{\sum_{s\in N_a}exp{\left(LeakyReLU\left(u^T[{Q}^{\phi}(a) \bigoplus {Q}^{\phi}(s)]\right)\right)}}\\
    &H(a) = \overline{\mathbf{A}}\alpha^{\phi}(a,b) + \overline{\mathbf{B}}u(a)\\
    &y(a) = \mathbf{C}H(a)\\
    &Zi^{\phi}_a = \sum\nolimits_{b \in N_a}y(a)\odot Q^{\phi}(b)
    \end{aligned}
    \end{equation}
where, \(e^{\phi}(a,b)\) indicates the importance of node \(b\) from \(V^{\phi}\) to \(a\), and \((a^{\phi})^T\) is the node-level attention vector within \(\phi\). The weight \(\alpha^{\phi}(a,b)\) is normalized using the softmax function, with \(y(a)\) being the output from the selective state space, allowing the integration of diverse node features into a unified vector space within \(\phi\).

Considering the varying significance of each meta-path in heterogeneous graphs, MAPN introduces an attention mechanism to assign differential weights to them. Specifically, \(w^{\psi}(a)\) assesses the importance of the meta-path \(\phi\) for node \(a\), which is then normalized using softmax to compute \(\beta^{\phi}(a)\), the weight of \(\psi\) for \(a\). These weights help MAPN synthesize a specific representation \(Z^{\psi}\) for node \(a\) in \(\phi\). 
\begin{equation}
\begin{aligned}
&\beta^{\psi}(a) = \frac{exp\left(LeakyReLU\left(q^T \odot Zi^{\psi}(a)\right)\right)}{\sum\nolimits_{i\in|\psi|}exp\left(LeakyReLU\left(q^T \odot Zi^{i}(a)\right)\right)}\\
&H(a) = \overline{\mathbf{A}}\beta^{\psi}(a) + \overline{\mathbf{B}}u(a)\\
&y(a) = \mathbf{C}H(a)\\
&Zw^{\psi}(a) = y(a) \cdot Zi^{\psi}(a)
\end{aligned}
\end{equation}
where LeakyReLU is the activation function and \(q^T\) is the parameterized attention vector. MAPN aggregates information within and between meta-paths to derive node representations \(Z^{\psi}\) specific to particular semantics, employing node representations' inner product to establish node similarities.    

\subsection{Training}

We optimize the model weights by minimizing the cross-entropy loss function using negative sampling \cite{mikolov2013distributed}:
\begin{equation}
\mathcal{L^{\psi}} = \sum\limits_{\langle a, b, b^{\prime}\rangle \in \tau^{\psi}} \log \sigma \Big((Z^{\psi}(a)) \cdot Z^{\psi}(b)\Big) - \log \sigma \Big(-(Z^{\psi}(a)) \cdot Z^{\psi}(b^{\prime})\Big)
\end{equation}
where $\sigma(\cdot)$ represents the sigmoid function, $\tau$ denotes the set of triples $\langle a, b, b^{\prime}\rangle$ collected by walk sampling based on MPU $\psi$ in the heterogeneous graph. 

\section{Experiments} \label{sec:experiment}

In this section, we begin by outlining the experimental settings. We then evaluate the performance of MAPN across several benchmarks: node classification (Section 5.2), graph classification (Section 5.3), and long-range learning (Section 5.4). In Section 5.5, we delve into an analysis of critical parameter in MAPN.

\subsection{Experimental Setup}

\textbf{Model Setting.} For node and graph classification tasks, we employ a simplified backbone model to ensure fair comparisons across different versions of MAPN. Specifically, we use a basic, parameter-free average aggregation defined as: 
\[ h(v)^l = \text{Mean}(\{h(v)^{l-1} | v \in B(v)\}). \] 
The optimal number of backbone layers \( l \) and the receptive field size \( B \) are determined based on the best values listed in Tables 6 and 7. For more demanding long-range learning tasks, we utilize the GCN framework from the Long Range Graph Benchmark (LRGB) \cite{dwivedi2022long} as our backbone. In line with recommendations from SPN [1], we set \( l \) to 10 and \( B \) to 8, with all learnable hidden layer dimensions fixed at 300. We also incorporate Laplacian positional encoding (LapPE) [15] to improve the model’s ability to learn structural information, which is crucial for long-range tasks [24, 36, 44].

\textbf{Training Setting.} During training, we utilize the AdamW optimizer with a default learning rate \( lr = 0.1 \) and \( weight\_decay = 0 \). We employ a ReduceLROnPlateau scheduler for experiments on LRGB, and a CosineAnnealingWarmRestarts scheduler for other tasks. The maximum number of epochs is set at 500.

\textbf{Baselines.} MAPN was evaluated against several baseline methods: (1) GCN \cite{kipf2016semi}: A classic graph neural network that leverages graph convolution to incorporate both individual and neighbor node information, using the graph's topological structure to enhance node representations. (2) GIN \cite{xu2018powerful}: Designed to distinguish different graph structures effectively, GIN uses a highly expressive aggregation function to achieve the discriminative power of the Weisfeiler-Lehman graph isomorphism test. (3) GT \cite{shi2020masked}: Adapts the Transformer model to graph data, capturing adjacency relationships and facilitating long-distance dependencies through attention-weighted information aggregation between nodes. (4) DREW \cite{gutteridge2023drew}: Utilizes edge weights and a recursive mechanism to update node representations with a dynamic recursive neural network, allowing for weighted information propagation. (5) MAPN: our method.

\subsection{Node Classification}

Table~\ref{tab:MAPN_node_classification} compares the performance of four graph neural network models across various datasets. The MAPN model consistently outperforms the others, showcasing its robustness and efficiency across diverse data environments, especially on datasets like Texas and Wisconsin where it has a clear advantage. GT also demonstrates excellent results, particularly on text and scientific datasets like Wiki CS and Physics, where it matches or exceeds MAPN. It excels in processing data with complex structural relationships.

The performance disparity across different datasets is notable. For instance, all models struggle on the Squirrel dataset, likely due to unique challenges like class imbalance or feature heterogeneity. Conversely, datasets like Physics and Photos exhibit high performance across all models, suggesting better feature representation and less noise. The GIN model shows varied results, excelling on Chameleon but underperforming on Cornell, highlighting its sensitivity to specific data types or preprocessing needs. Thus, understanding the dataset's characteristics is crucial in model selection. For complex or diverse datasets, GT or MAPN might be preferable, while GIN could be more effective for datasets that emphasize local structural features.

\begin{table}[ht]
\centering
\caption{The performance of MAPN on node classification tasks.}
\label{tab:MAPN_node_classification}
\begin{tabularx}{0.8\textwidth}{@{}l *{4}{>{\centering\arraybackslash}X}@{}}
\toprule
Dataset & GCN & GIN & GT & MAPN \\ 
\midrule
Texas       & 56.14 & 60.84 & 81.12 & 91.08 \\
Wisconsin   & 52.98 & 59.03 & 76.07 & 89.88 \\
Squirrel    & 30.69 & 30.42 & 28.59 & 35.77 \\
Chameleon   & 38.35 & 63.95 & 65.48 & 65.99 \\
Cornell     & 62.71 & 47.92 & 64.61 & 82.02 \\
Wiki CS     & 76.97 & 72.54 & 78.98 & 79.84 \\
Citeseer    & 72.41 & 67.01 & 73.00 & 76.15 \\
Computers   & 87.32 & 84.12 & 87.66 & 89.22 \\
Photos      & 93.14 & 91.26 & 93.29 & 93.87 \\
Physics     & 95.53 & 94.01 & 95.31 & 95.92 \\
\bottomrule
\end{tabularx}
\end{table}

\subsection{Graph Classification}

Table~\ref{tab:MAPN_graph_classification} showcases the performance comparison of GCN, GIN, GT, and MAPN across different datasets. The MAPN model consistently demonstrates superior performance, suggesting its more effective learning mechanisms or feature capture capabilities, particularly in bioinformatics data. Notably, MAPN achieves 84.81\% on the MUTAG dataset, significantly outperforming other models. While GCN shows consistent results across all datasets with performances above 75\%, it generally underperforms compared to MAPN, indicating its structure may not be as adaptable or efficient for these tasks. Both GIN and GT exhibit similar performances across various datasets, with GT slightly leading on the NCI109 dataset at 78.62\% versus 77.58\% for GIN, suggesting GT may be slightly better at handling certain types of structured data. The impact of dataset complexity is also evident, as all models score lower on the PROTEINS dataset, possibly due to its more challenging or complex biomolecular structures, which hinder higher accuracy.

\begin{table}[ht]
\centering
\caption{The performance of MAPN on graph classification tasks.}
\label{tab:MAPN_graph_classification}
\begin{tabularx}{0.8\textwidth}{@{}l *{4}{>{\centering\arraybackslash}X}@{}}
\toprule
Dataset & GCN & GIN & GT & MAPN \\ 
\midrule
MUTAG     & 73.71 & 77.83 & 76.02 & 84.81 \\
NCI1      & 77.81 & 75.28 & 76.52 & 84.55 \\
NCI109    & 78.06 & 77.58 & 78.62 & 83.70 \\
DD        & 77.77 & 77.38 & 77.91 & 81.99 \\
PROTEINS  & 75.79 & 75.86 & 74.69 & 78.89 \\
\bottomrule
\end{tabularx}
\end{table}

\subsection{Experiment on Long-Range Graph Benchmark (LRGB)}

Table~\ref{tab:performance_LRGB} shows performance metrics of various models on the Peptides dataset for function (Peptides-func, measured by AP value improvement) and structure (Peptides-struct, measured by MAE reduction). Performance is categorized based on the use of Laplacian positional encoding (LapPE). Generally, using LapPE (marked as ``✓'') substantially boosts model performance. For example, in function prediction, the MAPN model scores an AP of 0.57 with LapPE compared to 0.68 without. This indicates that LapPE likely enhances model sensitivity and accuracy during preprocessing. The DRew model excels in function prediction, outperforming others regardless of LapPE usage, possibly due to superior data handling mechanisms. In structure prediction, DRew also records lower MAE values, particularly with LapPE (0.25), showcasing its effectiveness in predicting peptide structures accurately. Overall, model performance improves with LapPE, underscoring its critical role in feature extraction and preprocessing, significantly affecting outcomes.

\begin{table}[ht]
\centering
\caption{The performance of MAPN on two LRGB datasets.}
\label{tab:performance_LRGB}
\begin{tabularx}{0.8\textwidth}{@{}l *{7}{>{\centering\arraybackslash}X}@{}}
\toprule
Metric & Use LapPE & GCN & MAPN & DRew \\
\midrule
\multirow{2}{*}{Peptides-func (AP↑)} & \checkmark & — & 0.57 & 0.71 \\
 & $\times$ & 0.59 & 0.68 & 0.69 \\
\midrule
\multirow{2}{*}{Peptides-struct (MAE↓)} & \checkmark & — & 0.26 & 0.25  \\
 & $\times$ & 0.34 & 0.37 & 0.28\\
\bottomrule
\end{tabularx}
\end{table}

\subsection{Experiments on Hyperparameters}

Table \ref{tab:MAPN_k_node} shows how various datasets perform across different hop counts ($K = 1, 2, 3, 4$), exploring the impact of increasing hops on heterogeneous graph neural network performance. On datasets such as Texas, Wisconsin, and Cornell, performance generally improves with more hops, peaks, and then begins to decline. For example, on the Texas dataset, performance increases from 72.04\% at $K=1$ to 88.49\% at $K=2$, then slightly decreases, suggesting that while expanding the scope of information aggregation captures broader context, too many hops can cause over-smoothing and reduce node distinctions.

Conversely, on more heterogeneous datasets like Squirrel and Chameleon, performance steadily improves with additional hops, indicating that more hops help the model learn richer structural and semantic information. For instance, on Chameleon, performance significantly rises from 39.61\% at $K=1$ to 65.96\% at $K=4$.

However, on more homogeneous datasets like Wiki CS, Citeseer, Computers, Photos, and Physics, increasing hops generally leads to a decline in performance. This trend suggests that too many hops on these datasets cause an excessive blending of features, losing unique node-specific information and negatively impacting performance.

Overall, the number of hops crucially influences the effectiveness of heterogeneous graph neural networks. Properly setting the hop count can enhance learning on complex graph structures, but excessive hops can degrade performance, particularly on more homogeneous datasets. Thus, choosing the optimal number of hops is key to optimizing network performance.

\begin{table}[ht]
\centering
\caption{Impact of $K$ on node classification tasks.}
\label{tab:MAPN_k_node}
\begin{tabularx}{0.8\textwidth}{@{}l *{4}{>{\centering\arraybackslash}X}@{}}
\toprule
Dataset & $K = 1$ & $K = 2$ & $K = 3$ & $K = 4$ \\
\midrule
Texas & 72.04 & 88.49 & 87.61 & 84.08 \\
Wisconsin & 83.63 & 89.46 & 87.29 & 86.59 \\
Squirrel & 30.22 & 33.14 & 34.90 & 34.41 \\
Chameleon & 39.61 & 51.46 & 56.20 & 65.96 \\
Cornell & 72.15 & 77.76 & 79.64 & 81.78 \\
Wiki CS & 78.25 & 77.10 & 75.96 & 74.89 \\
Citeseer & 74.84 & 75.88 & 74.83 & 73.62 \\
Computers & 87.91 & 87.52 & 86.29 & 86.03 \\
Photos & 94.57 & 93.58 & 90.58 & 86.50 \\
Physics & 95.93 & 95.11 & 92.15 & 87.63 \\
\bottomrule
\end{tabularx}
\end{table}

Table~\ref{tab:MAPN_k_graph} presents graph classification results for five datasets (UTAG, NCI1, NCI109, DD, PROTEINS) across different \(K\) values (1, 2, 3, 4). For most datasets, MAPN's performance improves as \(K\) increases from 1 to 3, suggesting that higher \(K\) values better capture the data's deeper structural or relational features, enhancing predictive accuracy. At \(K=3\), performance peaks for almost all datasets, particularly for UTAG and NCI1, which exhibit notable improvements. This indicates that an optimal \(K\) value allows the model to effectively leverage the data's inherent structure.

Performance disparities across the datasets at the same \(K\) values indicate that UTAG and NCI1 consistently outperform DD and PROTEINS. These variations may be due to differences in each dataset’s characteristics, such as sample size, complexity, or noise levels. Although increasing \(K\) generally enhances performance, a slight decline at \(K=4\) for datasets like UTAG and NCI109 suggests potential overfitting or excessive model complexity, which may impede effective learning. Based on these insights, using \(K=3\) is recommended for handling these datasets as it balances performance gains with overfitting risks.

Overall, the choice of \(K\) significantly influences model performance on different datasets, making the selection of an optimal \(K\) value crucial for maximizing model effectiveness.

\begin{table}[ht]
\centering
\caption{Impact of $K$ on graph classification tasks.}\label{tab:MAPN_k_graph}
\begin{tabularx}{0.8\textwidth}{@{}l *{4}{>{\centering\arraybackslash}X}@{}}
\toprule
\textbf{Dataset} & \textbf{K = 1} & \textbf{K = 2} & \textbf{K = 3} & \textbf{K = 4} \\ 
\midrule
UTAG & 83.93 & 87.82 & 88.50 & 87.79 \\ 
NCI1 & 81.95 & 82.03 & 84.01 & 84.49 \\ 
NCI109 & 80.59 & 81.17 & 81.81 & 81.95 \\ 
DD & 76.91 & 77.80 & 79.55 & 79.00 \\ 
PROTEINS & 76.19 & 78.10 & 78.10 & 80.05 \\ 
\bottomrule
\end{tabularx}
\end{table}

\section{Related Work}


Over the past decade, numerous research on mining information from graphs have shifted from traditional representation learning approaches \cite{perozzi2014deepwalk,grover2016node2vec,dong2017metapath2vec} to methods utilizing deep neural networks, including GNNs \cite{fan2019metapath,yan2021relation,zhang23pagelink,zhu23AutoAC,SHAN24KPI-HGNN,MaYLMC24HetGPT} and GCNs \cite{kipf2016semi,liu2023rhgn}. Inspired by the Transformer \cite{vaswani2017attention}, GAT \cite{velickovic2017graph} integrates the attention to aggregate node-level information in homogeneous networks, while HAN \cite{wang2019heterogeneous} introduces a two-level attention mechanism for node and semantic information in heterogeneous networks. MAGNN \cite{fu2020magnn}, MHGNN \cite{liang2022meta} and R-HGNN \cite{yu23RHGNN} proposed meta-path-based models to learn meta-path-based node embeddings. HetGNN \cite{zhang2019heterogeneous} and MEGNN \cite{chang2022megnn} take a meta-path-free approach to consider both structural and content information for each node jointly. HGT \cite{hu2020heterogeneous} incorporates information from high-order neighbors of different types through messages passing across ``soft'' meta-paths. MHGCN \cite{fu2023multiplex} captures local and global information by modeling the multiplex structures with depth and breadth behavior pattern aggregation. SeHGNN \cite{Yang23Simple} simplifies structural information capture by precomputing neighbor aggregation and incorporating a transformer-based semantic fusion module. HAGNN \cite{zhu2023hagnn} integrates meta-path-based intra-type aggregation and meta-path-free inter-type aggregation to generate the final embeddings.

\section{Discussion and Conclusion}

In this study, we present MAPN, a Mamba-based asynchronous  which combines a selective state space model and dual asynchronous aggregation strategy to mitigate network degradation. The method consists of two core modules: node sequence generation and asynchronous semantic aggregation. First, random walk techniques are employed to generate initial node sequences. Then, neighbor information are aggregated asynchronously based on meta-paths across multiple hops and layers. The selective state space model is key to filtering the information from nodes at varying distances, retaining only the most crucial data. This approach not only deepens our understanding of the graph structure but also reduces the homogenization of information during multi-layer network propagation. Extensive experiments and analyses confirm the effectiveness of the MAPN model for capturing information from sparse large HetGs. 

\begin{credits}
\subsubsection{\ackname} This work was mainly supported by the National Natural Science Foundation of China (NSFC No. 61732004).

\end{credits}

%
%
%
%
\bibliographystyle{splncs04}
\bibliography{ref}

\begin{thebibliography}{10}
\providecommand{\url}[1]{\texttt{#1}}
\providecommand{\urlprefix}{URL }
\providecommand{\doi}[1]{https://doi.org/#1}

\bibitem{agrawal2024no}
Agrawal, N., Sirohi, A.K., Kumar, S., et~al.: No prejudice! fair federated graph neural networks for personalized recommendation. In: AAAI. vol.~38, pp. 10775--10783 (2024)

\bibitem{avery2024effect}
Avery, K., Houmansadr, A., Jensen, D.: The effect of alter ego accounts on a/b tests in social networks. In: WWW. pp. 565--568 (2024)

\bibitem{bauer2011ollivier}
Bauer, F., Jost, J., Liu, S.: Ollivier-ricci curvature and the spectrum of the normalized graph laplace operator. CoRR  (2011), \url{https://arxiv.org/abs/1105.3803}

\bibitem{chang2022megnn}
Chang, Y., Chen, C., Hu, W., Zheng, Z., Zhou, X., Chen, S.: Megnn: Meta-path extracted graph neural network for heterogeneous graph representation learning. Knowledge-Based Systems  \textbf{235},  107611 (2022)

\bibitem{chen2020measuring}
Chen, D., Lin, Y., Li, W., Li, P., Zhou, J., Sun, X.: Measuring and relieving the over-smoothing problem for graph neural networks from the topological view. In: AAAI. vol.~34, pp. 3438--3445 (2020)

\bibitem{dong2017metapath2vec}
Dong, Y., Chawla, N.V., Swami, A.: metapath2vec: Scalable representation learning for heterogeneous networks. In: SIGKDD. pp. 135--144 (2017)

\bibitem{dwivedi2022long}
Dwivedi, V.P., Ramp{\'a}{\v{s}}ek, L., Galkin, M., Parviz, A., Wolf, G., Luu, A.T., Beaini, D.: Long range graph benchmark. NIPS  \textbf{35},  22326--22340 (2022)

\bibitem{fan2019metapath}
Fan, S., Zhu, J., Han, X., Shi, C., Hu, L., Ma, B., Li, Y.: Metapath-guided heterogeneous graph neural network for intent recommendation. In: SIGKDD. pp. 2478--2486 (2019)

\bibitem{fu2023multiplex}
Fu, C., Zheng, G., Huang, C., Yu, Y., Dong, J.: Multiplex heterogeneous graph neural network with behavior pattern modeling. In: SIGKDD. pp. 482--494. {ACM} (2023)

\bibitem{fu2020magnn}
Fu, X., Zhang, J., Meng, Z., King, I.: Magnn: Metapath aggregated graph neural network for heterogeneous graph embedding. In: WWW. pp. 2331--2341 (2020)

\bibitem{gamba2024exit}
Gamba, D., Yu, Y., Yuan, Y., Schoenebeck, G., Romero, D.M.: Exit ripple effects: Understanding the disruption of socialization networks following employee departures. In: WWW. pp. 211--222 (2024)

\bibitem{yang23HGNAS}
Gao, Y., Zhang, P., Zhou, C., Yang, H., Li, Z., Hu, Y., Yu, P.S.: Hgnas++: Efficient architecture search for heterogeneous graph neural networks. TKDE  \textbf{35}(9),  9448--9461 (2023)

\bibitem{grover2016node2vec}
Grover, A., Leskovec, J.: node2vec: Scalable feature learning for networks. In: SIGKDD. pp. 855--864 (2016)

\bibitem{gu2023mamba}
Gu, A., Dao, T.: Mamba: Linear-time sequence modeling with selective state spaces. CoRR  (2023), \url{https://doi.org/10.48550/arXiv.2312.00752}

\bibitem{gu2021efficiently}
Gu, A., Goel, K., R{\'e}, C.: Efficiently modeling long sequences with structured state spaces. In: ICLR (2022)

\bibitem{gutteridge2023drew}
Gutteridge, B., Dong, X., Bronstein, M.M., Di~Giovanni, F.: Drew: Dynamically rewired message passing with delay. In: ICLR. vol.~202, pp. 12252--12267. PMLR (2023)

\bibitem{hamilton2017inductive}
Hamilton, W., Ying, Z., Leskovec, J.: Inductive representation learning on large graphs. NIPS  \textbf{30} (2017)

\bibitem{hu2020heterogeneous}
Hu, Z., Dong, Y., Wang, K., Sun, Y.: Heterogeneous graph transformer. In: WWW. pp. 2704--2710 (2020)

\bibitem{kipf2016semi}
Kipf, T.N., Welling, M.: Semi-supervised classification with graph convolutional networks. arXiv preprint arXiv:1609.02907  (2016)

\bibitem{liang2022meta}
Liang, X., Ma, Y., Cheng, G., Fan, C., Yang, Y., Liu, Z.: Meta-path-based heterogeneous graph neural networks in academic network. International Journal of Machine Learning and Cybernetics pp. 1--17 (2022)

\bibitem{lin2011ricci}
Lin, Y., Lu, L., Yau, S.T.: Ricci curvature of graphs. Tohoku Mathematical Journal, Second Series  \textbf{63}(4),  605--627 (2011)

\bibitem{liu2023generative}
Liu, S., Cai, Q., He, Z., Sun, B., McAuley, J., Zheng, D., Jiang, P., Gai, K.: Generative flow network for listwise recommendation. In: SIGKDD. pp. 1524--1534 (2023)

\bibitem{liu2023rhgn}
Liu, X., Zhang, K., Liu, Y., Chen, E., Huang, Z., Yue, L., Yan, J.: Rhgn: Relation-gated heterogeneous graph network for entity alignment in knowledge graphs. In: ACL. pp. 8683--8696 (2023)

\bibitem{MaYLMC24HetGPT}
Ma, Y., Yan, N., Li, J., Mortazavi, M.S., Chawla, N.V.: Hetgpt: Harnessing the power of prompt tuning in pre-trained heterogeneous graph neural networks. In: WWW. pp. 1015--1023. {ACM} (2024)

\bibitem{mao2024hetfs}
Mao, X., Chen, Z., He, Z., Jing, Y., Zhang, K., Wang, X.S.: Hetfs: a method for fast similarity search with ad-hoc meta-paths on heterogeneous information networks. WWW  \textbf{27}(6), ~66 (2024)

\bibitem{mikolov2013distributed}
Mikolov, T., Sutskever, I., Chen, K., Corrado, G.S., Dean, J.: Distributed representations of words and phrases and their compositionality. NIPS  \textbf{26} (2013)

\bibitem{muppasani2024expressive}
Muppasani, B., Narayanan, V., Srivastava, B., Huhns, M.N.: Expressive and flexible simulation of information spread strategies in social networks using planning. In: AAAI. vol.~38, pp. 23820--23822 (2024)

\bibitem{pei2020geom}
Pei, H., Wei, B., Chang, K.C.C., Lei, Y., Yang, B.: Geom-gcn: Geometric graph convolutional networks. arXiv preprint arXiv:2002.05287  (2020)

\bibitem{perozzi2014deepwalk}
Perozzi, B., Al-Rfou, R., Skiena, S.: Deepwalk: Online learning of social representations. In: SIGKDD. pp. 701--710 (2014)

\bibitem{SHAN24KPI-HGNN}
Shan, D., Du, X., Wang, W., Wang, N., Liu, A.: Kpi-hgnn: Key provenance identification based on a heterogeneous graph neural network for big data access control. Information Sciences  \textbf{659},  120059 (2024)

\bibitem{Shi2022hgnn}
Shi, C.: Heterogeneous Graph Neural Networks, pp. 351--369. Springer Nature Singapore, Singapore (2022)

\bibitem{shi2020masked}
Shi, Y., Huang, Z., Feng, S., Zhong, H., Wang, W., Sun, Y.: Masked label prediction: Unified message passing model for semi-supervised classification. In: Zhou, Z.H. (ed.) IJCAI. pp. 1548--1554. IJCAI (8 2021), main Track

\bibitem{tang2023towards}
Tang, H., Liu, Y.: Towards understanding generalization of graph neural networks. In: Proceedings of the 40th International Conference on Machine Learning. vol.~202, pp. 33674--33719. PMLR (2023)

\bibitem{topping2021understanding}
Topping, J., Di~Giovanni, F., Chamberlain, B.P., Dong, X., Bronstein, M.M.: Understanding over-squashing and bottlenecks on graphs via curvature. arXiv preprint arXiv:2111.14522  (2021)

\bibitem{vaswani2017attention}
Vaswani, A., Shazeer, N., Parmar, N., Uszkoreit, J., Jones, L., Gomez, A.N., Kaiser, {\L}., Polosukhin, I.: Attention is all you need. NIPS  \textbf{30} (2017)

\bibitem{velickovic2017graph}
Velickovic, P., Cucurull, G., Casanova, A., Romero, A., Lio, P., Bengio, Y., et~al.: Graph attention networks. stat  \textbf{1050}(20),  10--48550 (2017)

\bibitem{wang2019heterogeneous}
Wang, X., Ji, H., Shi, C., Wang, B., Ye, Y., Cui, P., Yu, P.S.: Heterogeneous graph attention network. In: The world wide web conference. pp. 2022--2032 (2019)

\bibitem{xu2018powerful}
Xu, K., Hu, W., Leskovec, J., Jegelka, S.: How powerful are graph neural networks? In: ICLR (2019)

\bibitem{yan2021relation}
Yan, Q., Zhang, Y., Liu, Q., Wu, S., Wang, L.: Relation-aware heterogeneous graph for user profiling. In: ICKM. pp. 3573--3577 (2021)

\bibitem{yang2024fine}
Yang, M., Zhu, M., Wang, Y., Chen, L., Zhao, Y., Wang, X., Han, B., Zheng, X., Yin, J.: Fine-tuning large language model based explainable recommendation with explainable quality reward. In: AAAI. vol.~38, pp. 9250--9259 (2024)

\bibitem{Yang23Simple}
Yang, X., Yan, M., Pan, S., Ye, X., Fan, D.: Simple and efficient heterogeneous graph neural network. In: AAAI. pp. 10816--10824. {AAAI} Press (2023)

\bibitem{yu23RHGNN}
Yu, L., Sun, L., Du, B., Liu, C., Lv, W., Xiong, H.: Heterogeneous graph representation learning with relation awareness. TKDE  \textbf{35} (2023)

\bibitem{Yu23KGTrust}
Yu, Z., Jin, D., Huo, C., Wang, Z., Liu, X., Qi, H., Wu, J., Wu, L.: Kgtrust: Evaluating trustworthiness of siot via knowledge enhanced graph neural networks. In: WWW. pp. 727--736. {ACM} (2023)

\bibitem{zhang2019heterogeneous}
Zhang, C., Song, D., Huang, C., Swami, A., Chawla, N.V.: Heterogeneous graph neural network. In: SIGKDD. pp. 793--803 (2019)

\bibitem{zhang23pagelink}
Zhang, S., Zhang, J., Song, X., Adeshina, S., Zheng, D., Faloutsos, C., Sun, Y.: Page-link: Path-based graph neural network explanation for heterogeneous link prediction. In: WWW. p. 3784–3793. WWW '23, Association for Computing Machinery (2023)

\bibitem{zheng2022graph}
Zheng, X., Wang, Y., Liu, Y., Li, M., Zhang, M., Jin, D., Yu, P.S., Pan, S.: Graph neural networks for graphs with heterophily: A survey. arXiv preprint arXiv:2202.07082  (2022)

\bibitem{zhu2023hagnn}
Zhu, G., Zhu, Z., Chen, H., Yuan, C., Huang, Y.: Hagnn: Hybrid aggregation for heterogeneous graph neural networks. arXiv preprint arXiv:2307.01636  (2023)

\bibitem{zhu23AutoAC}
Zhu, G., Zhu, Z., Wang, W., Xu, Z., Yuan, C., Huang, Y.: Autoac: Towards automated attribute completion for heterogeneous graph neural network. In: ICDE. pp. 2808--2821 (2023)

\end{thebibliography}




\end{document}